\documentclass[sigconf]{acmart}
\usepackage{booktabs} 
\usepackage{epsfig}
\usepackage{epstopdf}
\usepackage{amssymb,amsmath}
\usepackage{amsmath}
\usepackage{graphicx}
\usepackage{caption}
\usepackage{graphicx}

\setcopyright{rightsretained}
\acmConference{ICTIR' 17 Workshop on Search-Oriented Conversational AI (SCAI' 2017)}{October 1, 2017}{Amsterdam, Netherlands}

\begin{document}
\title{Perspectives for Evaluating Conversational AI}

\author{Mahipal Jadeja}
\affiliation{%
  \institution{DA-IICT}
  \city{Gandhinagar} 
  \state{India} 
}
\email{mahipaljadeja5@gmail.com}

\author{Neelanshi Varia}
\affiliation{%
 \institution{DA-IICT}
 \city{Gandhinagar} 
 \state{India} 
}
\email{neelanshiV2@gmail.com}

\begin{abstract}
Conversational AI systems are becoming famous in day to day lives. 
In this paper, we are trying to address the following key question: To identify whether design, as well as development efforts for search oriented conversational AI are successful or not. It is tricky to define 'success' in the case of conversational AI and equally tricky part is to use appropriate metrics for the evaluation of conversational AI. We propose four different perspectives namely user experience, information retrieval, linguistic and artificial intelligence for the evaluation of conversational AI systems.
Additionally, background details of conversational AI systems are provided including desirable characteristics of personal assistants, differences between chatbot and an AI based personal assistant. An importance of personalization and how it can be achieved is explained in detail. Current challenges in the development of an ideal conversational AI (personal assistant) are also highlighted along with guidelines for achieving personalized experience for users. 
\end{abstract}

%
%

\maketitle

\section{Introduction}
Science fiction movies and novels have numerous kinds of robots and computers and it was once a dream. But today, since past few years it is no more a dream. The artificial intelligence upsurge allowed us to talk to computers via commands. And conversational artificial intelligence has allowed us not just to talk to machines but also accomplish our tasks. Siri (Apple), Google Now (Google), Cortana (Microsoft) and Alexa (Amazon) have revolutionized the way we perceive phones and machines. Some features of these personal assistants are shown in Figure~\ref{AIf}. These assistants are  termed as "dialogue systems often endowed with humanlike behaviour" $[1]$, they have started becoming integral parts of people's lives. In the coming years, a man will truly have an artificial friend who will talk, do work and give suggestions just like a human friend would have done. There are immense problems lying in the field of conversational AI to be solved to reach the dream we pursue. Techniques such as Natural Language Processing, Machine Learning, Artificial Neural Networks, etc. come to our aid. We have clearly moved ahead of normal conversation based text chatbots but now is the time to increase the efficiency of speech-based and predictive, artificially intelligent assistants. PAs which can solve really deep and complex conversations and can talk as human-like as possible.

\subsection{Difference between a chatbot and an artificially intelligent PA:}
\noindent Chatbot: Chatbots are usually programs which are meant to have conversations with human users via text or speech methods. They are at times meant for specific tasks in various companies and sometimes for general chit-chat purposes. They are subset or parts of AI bots/assistants rather than being complete Virtual assistants. Example - FaceBook Messenger bot, Google Allo, etc. Chatbots aren't supposed to perform tasks apart from having a conversation and providing relevant information.\\

\begin{figure}[h]
	\includegraphics[scale=0.2]{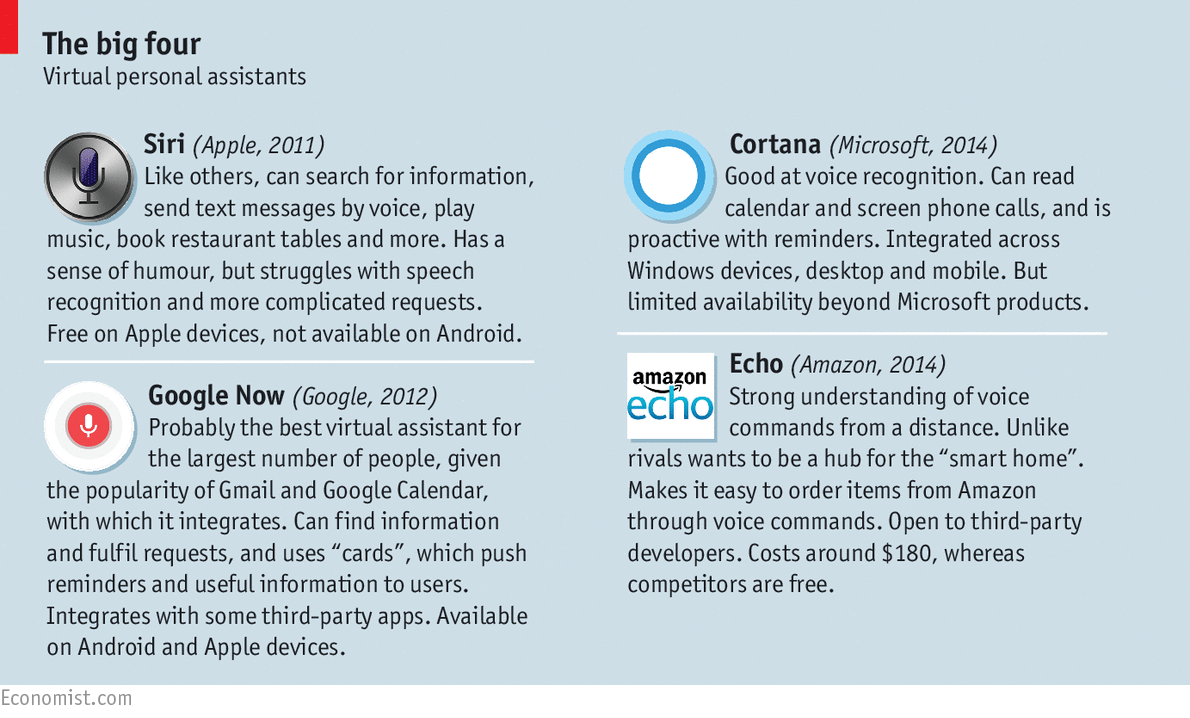}
	\caption{Some features of well-known virtual personal assistants}
	\label{AIf}
\end{figure}

\begin{figure}[h]
	\includegraphics[scale=0.35]{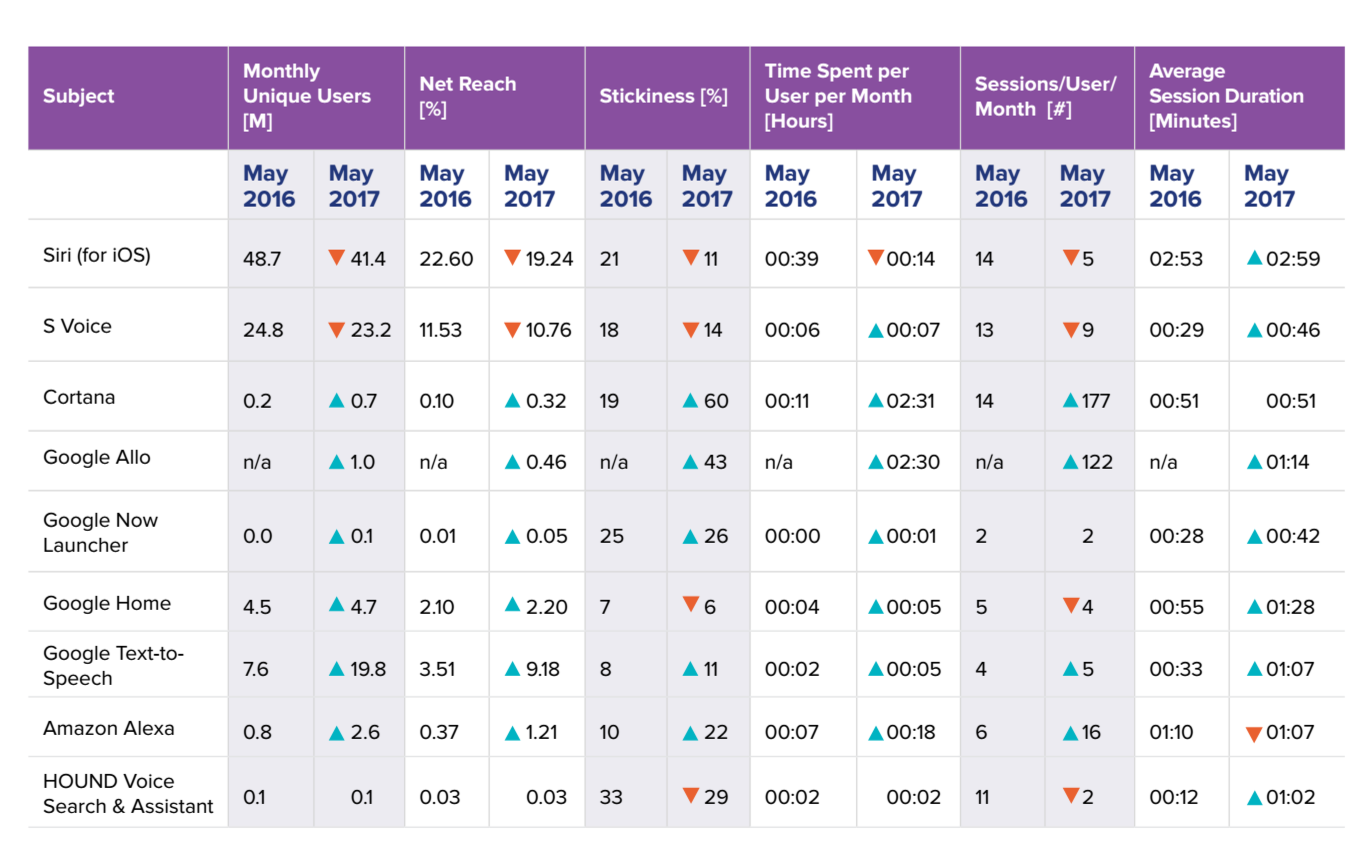}
	\caption{Usage statistics of well-known virtual personal assistants}
	\label{pqr}
\end{figure}

\noindent Artificial assistant: The virtual intelligent assistants are not just ones who perform conversations but they perform tasks. They are also built up on complex algorithms of Natural Language Processing, Machine Learning, and Artificial Neural Networks. It learns along with its usage and gives better performance whereas the chatbots are based on some rules and regulations which further are not modified.  Figure~\ref{AIf} explains features and limitations of four famous virtual personal assistants. Usage statistics of various virtual personal assistants are shown in Figure~\ref{pqr}.

\subsection{Characteristics of a PA }
In this subsection, we list key characteristics of a human personal assistant and we want all these characteristics in virtual personal assistant (AI based) in future.

\textbf {Knowing the expectations of the user:}
It is important to know preferences of the user like timings of the meeting, how to attend other employees, tea-coffee preference, way of keeping MoM, etc. It is also important to know the peculiarity of the user so as to react accordingly.

\textbf {Efficient organisation:}
The PA should have the files, documents, etc. organised in an efficient and handy manner. Be it in an electronic database or hard copies in the office. Various contacts, email addresses, plans, presentations, etc. should be in near reach.

\textbf {Security and trust:}
One of the most important aspects is not breaching trust and maintaining the secrecy of information provided by user. Internal statistics of the company, private contacts and future plans, track record and schedules which are confidential should not be leaked anywhere else.

\textbf{Understanding  tactics of users in his/her job:}
It is important to know the methodology of the business/company, various rules, etc. in order to converse with other professionals and employees. The PA should thoroughly understand the way the company works and its strategies.

\textbf{Making a perfect schedule:}
According to the order of phases of any project, priorities of the user and other employees, temporal preferences, past experiences, etc., the PA should be able to draft a perfect schedule. 

\textbf{Problem solving and influencing skills:}
One of the most important quality of a human PA is that he/she understands the persona of the boss and try to present the boss with circumstance dependent solution. During stressful or dilemmatic situation, the PA should be able to come up with stress relieving words or alternatives based on his/her own intellect.

\section{Evaluation of search-oriented conversational AI}
\textbf {Key question:} To identify whether design, as well as development efforts for search oriented conversational AI are successful or not.\\ \\
If a conversational AI is used for business purposes then the answer is quite simple:  Modification of the existing metrics and/or KPIs (Key Performance Indicator) in order to reflect properties like user engagement, retention etc.  But unfortunately, the same strategy can't be used for all conversational AIs since not all of them are used for business purposes. We need to identify a single universal mechanism to come up with metrics which can evaluate any conversational AI and we must ensure iterative implementation (feedback loop) which can be beneficial for improving the overall design.

We propose four different perspectives regarding conversational AIs. These perspectives can be useful in the design of metrics for conversational AIs. 

\begin{enumerate}
	\item \textbf{User Perspective:} Conversational AIs are designed for human-like interaction with end users but it seems very difficult to achieve this objective practically $[5]$. Key desirable features for virtual personal assistants are discussed in the introduction section. So natural way to evaluate AI based personal assistants is to consider parameters like usability, level of user satisfaction etc.  The natural measure is to compare the feasibility of doing a set of tasks in two different cases: 1) with the help of conversational AI 2) without the help of conversational AI. The user may take help of another human personal assistant in this case.  According to us, this is the most critical perspective. The main drawback is: this approach is expensive (in terms of time as well as money). Additionally, it is not practical to implement this approach on a large scale. 
	
	It is possible to improve following characteristics of a PA on successful execution  of user perspective evaluation metrics: a) To know the expectations of the user b) Establishment of security and trust by identification of private/confidential user data c) Understanding tactics of users. 
	
	\item \textbf{IR perspective:} It is possible to get a lot of quantitative data for measurement/evaluation of the conversational AI using this perspective. For example, the accuracy of information and how quickly the system reacts to the users' query. But good information retrieval (IR) qualities don't necessarily make users happy because user experiences depend upon various parameters. For example, a user wants to buy a wallet and the conversational AI replies very quickly by showing all the options (weblinks) for buying a wallet at the cheapest rate but still the user is unhappy because he/she wants to buy a branded wallet with a good return policy. 
	
	It is possible to design an efficient personal assistant by considering  IR perspectives for evaluation. 
	
	\item \textbf {Linguistic Perspective:} According to Grice $[2]$, a cooperative principle is shared between speakers and hearers during an ordinary conversation. He has identified 4 maxims in cooperation namely:
	\begin{itemize}
		\item Quality:  whatever is said by the speaker is truth and it can be proved with the help of evidence.
		\item 	Quantity:  the amount of information shared by the speaker must be depending upon the requirement. The speaker must not share too much or too less information.
		\item 	Relation: the response must be related to the discussion topic.
		\item 	Manner: no ambiguity/obscurity and the overall interaction must be direct as well as straightforward.
	\end{itemize}
	Grice's cooperative maxims and scoring matrix can also serve as a tool for comparing different conversational AIs $[3][4]$. So a good conversational AI will have a higher degree of support towards Grice's conversational maxims.  The good thing about this approach is that it considers critical elements for effective conversation but the drawback is it is dependent on expert judgement  (for example, who will decide whether the conversational AI's response is related to the topic or not?-relation maxim). Due to this drawback, it is difficult to use this scheme at large scale. 
	
	This approach is useful for improving trust between conversational AI and the user. 
	\item  \textbf {Artificial Intelligence Perspective:}  We may use Turing test $[6]$ types of measures in order to identify the presence/absence of human like interaction abilities.  But the problem with this approach is: it won't provide guidelines for the overall improvement of the conversational AI.  Let us say, we have developed one new conversational AI and it is failed in Turing test. Next is what?
	
	By analyzing artificial intelligent perspective based evaluation schemes, it is possible to improve problem solving and influencing skills of personal assistant. 
\end{enumerate}
According to us, IR approaches are more suitable for quantitative type of evaluation whereas user experience approaches are more suitable for qualitative type of evaluation.  In order to design a universal evaluation framework, hybrid evaluation is desirable.

Most of the new conversational based AIs are proactive search systems. Unlike traditional reactive services (where the behaviour is query then response type), push notifications are sent to users based on location, user interests, environment, time etc in the case of proactive search systems - zero query IR systems.  Due to these differences, special types of metrics must be designed for evaluation.  According to $[8]$, we should use manual (human) evaluation along with measures that only roughly approximate human judgements.

Sometimes it is beneficial to have a game like interactive style in conversational AIs. In this context, various known games UX techniques and evaluation metrics are more appropriate $[7]$.

Task Completion Platform is a system that supports multi-tasking using Language Units $[13]$. Tasks are defined separately from the dialogue policy that is each of them can be differently added or removed without requiring the system to rebuild. Further, we will have to work on the tasks like concurrently allowing them to process results. Each task here is an independent routine which begins after a set of parameters that are defined. Experience based learning is another feature like other existing mechanisms but it is different in the sense that it builds up a bigger experience by weaving smaller experiences. The personal digital assistant here behave like an application which fulfill the requirements after an input series is completed.

There are various end to end task management systems $[14]$ that use human judges. Depending on the input and system output, the response is rated from 1 to 5 and then scores are mapped in a success matrix. In such a system there is a Language Unit that allows system to correct mistakes with the help of features extracted from dialog and experience providers.

\section{Personalization}
An efficient AI based assistant should be able to understand the way of working of a user, his/her timings, reactions to various situations and predilections and then adapt accordingly. Client should be able to add various range of personalised tasks, their methods and timings. For such users the intelligent PA should also be able to memorize past and future meetings of the client and according to the commitments of the client at various places and tasks, the PA should schedule perfect meeting hours for the user using various ML, search-based AI and Deep Learning algorithms. 

\begin{figure}[h]
	\includegraphics[scale=2]{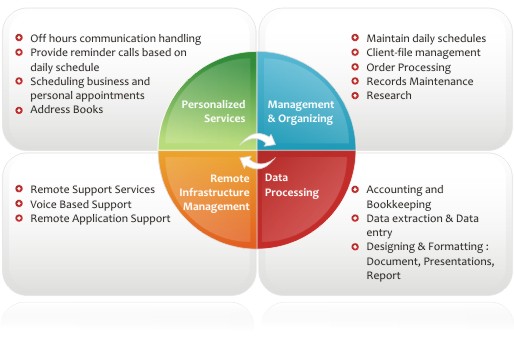}
	\caption{Personalization for users}
	\label{per4}
\end{figure} 

There are four major tasks which an artificial assistant is required to perform. Personalized services - management and organization of tasks, data storage and processing and remote infrastructure management. The latter three activities help towards giving personal experience to the users (see Figure~\ref{per4}).

While asked to schedule a meeting, the scheduling feature of the PA should be able to find appropriate slots according to the pre-defined preferences of the user and then user should be able to choose from them. The AI should also be able to provide the immediately preceding and succeeding meetings of the client instantly just in case if it is required. By various reinforcement learning algorithms and artificial neural networks, the PA should be able to learn the patterns and conditions of appointing a meeting/schedule. This scheduling feature earlier was only able to perform some common tasks by looking at date and time, etc.$[10]$ but now with the Machine Learning techniques at hand and various algorithms available, the PA should be able to learn on the basis of multiple features like scheduling based on the availability of meeting attendees, prioritize preferences of the client and other attendees, weather, commuting, etc. These constraints when assessed by the intelligent PA leads to almost perfect scheduling for the client and relieving them of mental calculations to be performed just for setting up a meeting allowing them to focus on the actual work. 

Active learning $[11]$ can be used for our desired purpose. But here only learning from the client's activity is the idea and hence there's no limitation to various schedule preferences of the clients. So here, using the active learning algorithm and using the above mentioned scheduling feature real-time perfect schedule settings can be obtained. Generally, entropy based criteria is used by the active learner in order to identify a set of feasible solutions $[12]$, to find optimum settings based on various possible meeting slots, constraints and requirements for the client.
It was seen from various results of this learning method that initially while the samples are still being trained the users come to a conclusion that AI isn't intelligent or upto their mark because the intelligent AI still needs more time to learn. So, what we can include in the scheduling feature is an initial Q and A based or instructions based learning period for the PA where users can set some initial conditions, priorities and instruct the PA to behave in a particular manner in a specific situation or time. Using these training instructions and iterations of learning over a period of time along with active learning desirable meeting schedules can be generated leading the client to satiety.

\subsection{How to make your personal assistant more 'personal'?}
After the development of an intelligent personal assistant with greater conversational and better prediction capabilities we need more of personalisation for the user. The assistants such as Ok Google, Siri, Cortana, Alexa etc. to much extent succeed in knowing your background and giving you a personalised experience, but are their characters a persona you want? Following are some characteristics we have a need to add in future to an intelligent assistant.

\textbf {Nomenclature and gender decision:} 
Cortana, Siri, Alexa, etc. are personal assistants round the world. To make the assistant really personal one should get an option to change the name. Also, by default all the assistants have female voice and that is majorly based on an assumption that both men and women find the voice of female better than male $[9]$. One should have the choice of selecting gender for his/her personal assistant. The PA can also be genderless.

\textbf{Language:}
Most of the assistants today have their default language English and a very few languages are available to switch into. A regional language and various accents of English should be available for the user so that he/she does not have to strain while listening or speaking. Many countries round the world don't speak English and globalising the assistants via this method would be a marvelous achievement.

\textbf {Trust:} 
The assistants have a lot of knowledge and discernment of the users. To give a personal experience to the users, the users should be able to put trust in the PAs in concern with personal delicate information and history of the user.

\textbf{Learning from the user:}
To be able to suggest users with trip plans, shopping suggestions, meetings, etc. based on past history of browsing and past answers, the PA should be able to assist with appropriate solutions. It should know various preferences of contacts, timings to do certain things, etc.

\textbf{Age and occupation of the user:}
The intelligent AI should be age adaptive. For instance, if a child is free and wants to do something the AI should provide him/her with various fun learning activities or games.\\
For a businessman it should be able to get some stock results or various happenings in his area.\\
For a housewife or anyone free at home, it should suggest various 'relevant' activities. Like in the following example, it can be seen that the VPA gives interesting options to the user.\\
\textit{ 
	Housewife: Hi Jenny! I am getting bored. Suggest something to do.\\
	Jenny (VPA): Hey Lisa! Here are some quick recipes I've found for you to make. Or you can try going to Square Garden. There's a fun fair going on and it should take half an hour for you to see the things around.}

\section{Future of Conversational AI}
With Siri, Cortana, Alexa and OK Google already in the market, some of the aspects need new dimensions. The Virtual Personal Assistant technology has applications in smart education, smart hygiene and healthcare, smart refrigerators, washing machines, TV, etc.

The personal assistants have been able to move one step ahead from 'entering commands' to recognising 'speech commands' using Natural Language Processing. But a lot more is needed in this area to allow its full-fledged use for the customers. These PAs could initially recognise only certain commands and words but the mammoth task of identifying natural language completely is an important. 
The challenges faced initially was the provision of vocabulary to the personal assistants, practical language recognition based on local slangs and accents. Understanding of context and reference based on history and previous data apart from natural language is a challenge we have to face. At many companies and institutes speech-based AI is the major focus for research to give users best performance and experience of AI. Also, research in AI for defense and AI for solving complex task is prime.

\begin{figure*}[h]
	\includegraphics[scale=0.39]{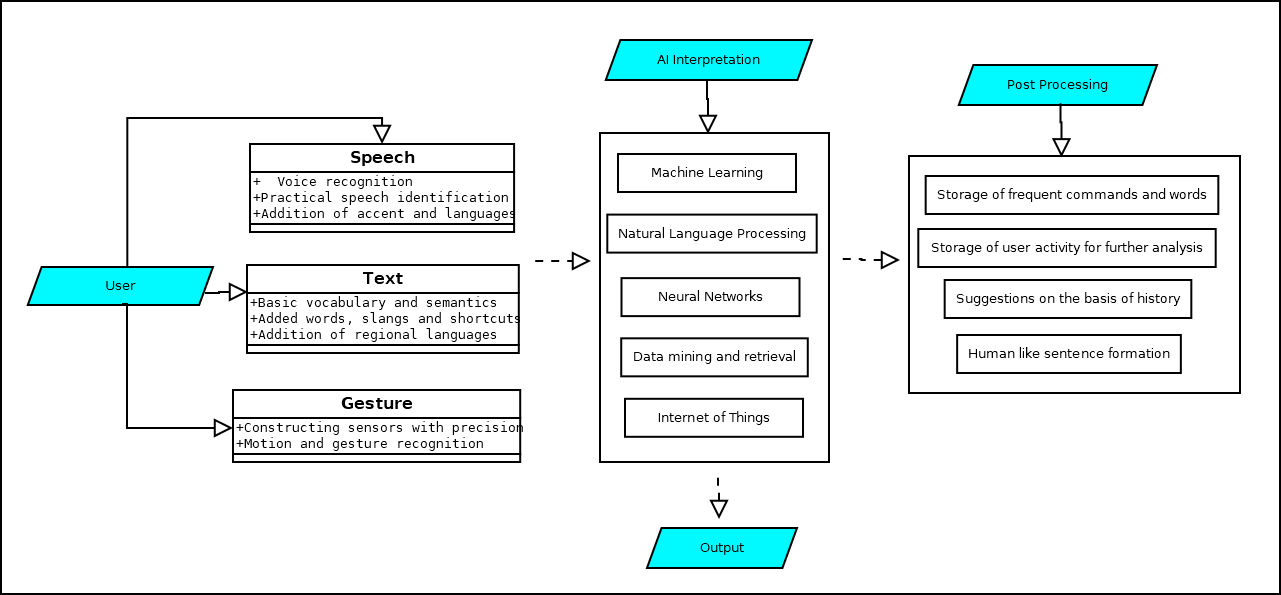}
	\caption{Advanced aspects of conversational AI}
	\label{adas}
\end{figure*} 

The future era of virtual intelligent assistants will allow us to have conversations in depth with the PAs. It will be like having a human relationship with them. The conversations will not just be complex but also be long-lasting, more suggestive, clear, with various emotions, added functionalities and tasks and most importantly, more personalised. It will be continuously finding user's likings and dislikings and adjust accordingly.

As mentioned above, we already have voice recognition into practice and basic gesture recognition. What we need is better post-data processing techniques in order to learn, remember and predict for the users. This can be accomplished with the help of large data storage capacities and optimum algorithms of the subjects mentioned in the Figure~\ref{adas}. Following is a brief explanation of the features depicted in Figure~\ref{adas}. The post processing stage requires maximum work as it fosters the user experience for the next time a user uses the device again.  

Storage of frequent commands and words and storage of activities: It is important to be faster and adaptive for an intelligent assistant and hence predicting further actions of users makes the experience smooth. Activities performed by users at a particular time of the day, week, etc , the context of the activity is important to predict future actions and recognise the pattern of different applications and commands given by user.

Suggestions based on history: After above two steps, most perfect prediction possible should be taken care of. 

Human like sentence formation: Apart from the basic grammar and structure, the intelligent assistant should be able to include various expression-based sentences and continuable conversations unlike the present precise and crisp answers.

Following is an example of a friendly, human-like, suggestive conversation between AI and a user :\\ 
\textit{
	User: Hey! I have my sister's wedding on $15^{th}$ September in New York.\\
	VPA: Wow, that's good to hear. You will have to book your tickets and bus journey doesn't seem to suit you well. Shall I look for a flight ticket?\\
	User: Oh yes, thank you. For five days.\\
	VPA: xyz airlines are having the cheapest rate right now but the timings are $3$ am in the morning while going. abc airlines cost only $20$ USD more but they have good seats and timings. What say?\\
	User: Right. What's the cost for abc?\\
	VPA: $100$ USD.\\
	User: Book them.\\
	VPA: Done!\\}
This conversation was based on history of user's travel, credit card and other stored details and various suggestions. The VPA itself suggests to book tickets, go by flights and sees the convenience of the timings. VPA has learnt things in past.

\section{Conclusions}
All the critical and relevant issues related to evaluation of conversational AI based systems (personal assistants) are discussed in this paper. Four different approaches for evaluating personal assistants are explained in detail. Importance of personalization and road-map for achieving it is also explained along with examples. Evaluation of personal assistants depends upon quantitative as well as qualitative features and most of the qualitative features are user dependent. Therefore, we feel that extreme personalization and universal defined metrics for qualitative features are the biggest challenges for conversational AI based personal assistants. 
\section{References}
\begin{enumerate}
	\item Vassallo, G., Pilato, G., Augello, A., \& Gaglio, S. $(2010)$. Phase Coherence in Conceptual Spaces for Conversational Agents (pp. $357-371$). John Wiley \& Sons, Inc., Hoboken, NJ, USA.
	\item Grice, H. P., Cole, P., \& Morgan, J. $(1975)$. Logic and conversation. $1975$, $41-58$.
	\item Vogel, A., Bodoia, M., Potts, C., \& Jurafsky, D. $(2013)$. Emergence of Gricean Maxims from Multi-Agent Decision Theory. In HLT-NAACL (pp. $1072-1081$).
	\item Chakrabarti, C., \& Luger, G. F. ($2013$, May). A Framework for Simulating and Evaluating Artificial Chatter Bot Conversations. In FLAIRS Conference.
	\item Luger, E., \& Sellen, A. ($2016$, May). Like having a really bad PA: the gulf between user expectation and experience of conversational agents. In Proceedings of the $2016$ CHI Conference on Human Factors in Computing Systems (pp.$ 5286-5297$). ACM.
	\item Warwick, K., \& Shah, H. ($2016$). The importance of a human viewpoint on computer natural language capabilities: a Turing test perspective. AI \& society, $31(2)$, $207-221$.
	\item Nacke, L. E. ($2015$). Games user research and physiological game evaluation. In Game user experience evaluation (pp. $63-86$). Springer International Publishing.
	\item Liu, C. W., Lowe, R., Serban, I. V., Noseworthy, M., Charlin, L., \& Pineau, J. $(2016)$. How NOT to evaluate your dialogue system: An empirical study of unsupervised evaluation metrics for dialogue response generation. arXiv preprint arXiv:$1603.08023.$
	\item Mitchell, W. J., Ho, C. C., Patel, H., \& MacDorman, K. F. ($2011$). Does social desirability bias favor humans? Explicit-implicit evaluations of synthesized speech support a new HCI model of impression management. Computers in Human Behavior, $27(1), 402-412.$
	\item Gervasio, M. T., Moffitt, M. D., Pollack, M. E., Taylor, J. M., \& Uribe, T. E. ($2005$, January). Active preference learning for personalized calendar scheduling assistance. In Proceedings of the 10th international conference on Intelligent user interfaces (pp. $90-97$). ACM.
	\item Cohn, D., Atlas, L., \& Ladner, R. $(1994)$. Improving generalization with active learning. Machine learning, $15(2), 201-221$.
	\item Weber, J. S., \& Pollack, M. E. ($2007$, January). Entropy-driven online active learning for interactive calendar management. In Proceedings of the $12^{th}$ international conference on Intelligent user interfaces (pp. $141-150$). ACM.
	\item Crook, P. A., Marin, A., Agarwal, V., Aggarwal, K., Anastasakos, T., Bikkula, R., \& Holenstein, R. (2016). Task Completion Platform: A self-serve multi-domain goal oriented dialogue platform.
	\item Sarikaya, R., Crook, P. A., Marin, A., Jeong, M., Robichaud, J. P., Celikyilmaz, A., \& Boies, D. (2016, December). An overview of end-to-end language understanding and dialog management for personal digital assistants. In Spoken Language Technology Workshop (SLT), 2016 IEEE (pp. 391-397). IEEE.
	
\end{enumerate}

\end{document}